\documentclass[submission,copyright,creativecommons]{eptcs}
\providecommand{\event}{Workshop on Agents and Robots for Reliable\\ Engineered Autonomy, ECAI 2020} 
\usepackage{breakurl}             
\usepackage{underscore}           
\usepackage{graphicx}
\usepackage{subfigure}
\usepackage{cite}
\usepackage{amsmath,amssymb,amsfonts}

\title{Establishing Reliable Robot Behavior using Capability Analysis Tables}
\author{Victoria Edwards \qquad\qquad Loy McGuire
  \institute{Naval Research Laboratory, Washington DC,  USA}
  \institute{Distributed Autonomous Systems Section, Code 5514}
  \email{\quad victoria.edwards@nrl.navy.mil  \quad\qquad  loy.mcguire@nrl.navy.mil}
  \and
  Signe Redfield
  \institute{Naval Research Laboratory, Washington DC, USA}\
  \institute{Robotics and Machine Learning Section, Code 8234 }
  \email{signe.redfield@nrl.navy.mil}
}
\def\titlerunning{Capability Analysis Tables }
\def\authorrunning{Victoria Edwards, Loy McGuire \& Signe Redfield}

\begin{document}

\maketitle
\begin{abstract}
Robots are often so complex that one person may not know all the ins and outs of the system. 
Inheriting software and hardware infrastructure with limited documentation and/or practical robot experience presents a costly challenge for an engineer or researcher. 
The choice is to either re-build existing systems, or invest in learning the existing framework. 
No matter the choice, a reliable system which produces expected outcomes is necessary, and while rebuilding may at first appear easier than learning the system, future users will be faced with the same choice. 
This paper provides a method to allow for increased documentation of the robotic system, which in turn can be used to contribute in overall robot reliability.
To do this we propose the identification of a robot's core behaviors for use in Capability Analysis Tables (CATs). 
CATs are a form of tabular documentation that connect the hardware and software inputs and outputs to the robot's core behaviors. 
Unlike existing methods, CATs are flexible, easy to build, and understandable by non-expert robot users. 
We demonstrate this documentation method with an experimental example using an Unmanned Aerial Vehicle (UAV). 
\end{abstract}

\section{Introduction}
Autonomous systems can have many hardware components and countless lines of code.
It  takes time to work through all the raw information about the systems' inner-workings, and in many instances, documentation is limited.
Despite this complexity, it has been demonstrated that some robots are able to function reliably in certain settings, but only when the environment is simple, or the level of autonomy is reduced.
Reliable, for the purposes of this paper, means that the output of the robot meets the expectations of what the robot should be doing consistently.
Improved reliability, is the identification or elimination of undefined output response for a provided input to the robot. 
When handling a new robotic system, it is easy to pass an input to a robot that could produce an unexpected behavior, especially with little documentation or intuition for what the robot should do.

One way to transfer knowledge to a new user is to spend time with the developer or former caretaker of the robot to learn from that person's experience.
Often, this does not or cannot happen. 
Because of the increasing desire for reliable robotic systems, it has become necessary to develop tools to better facilitate these types of transitions. 
There are many existing tools to ensure reliability of a robotic system, for example, finite state machines (FSMs)\cite{Chow78}, formal methods \cite{Lyons89}, Domain Specific Language (DSL) \cite{Nordmann14}, and Systems Modeling Language (SysML) \cite{powerpoint06}.
However, each of these tools require expert knowledge of either the robotic system or the methodology to generate useful representations of the system.

Tools and methods need to be developed to enable easier transfer of knowledge about existing robot capabilities to new users which leverage existing understanding of how the robot works. 
A step towards achieving this is to augment documentation to capture not only the software components but also how that software integrates with hardware.
First, it is necessary to identify the core behavior as a way to establish a performance baseline for the robotic system. 
\textbf{Core behaviors} are the high level description of what the robot should be able to do reliably with it's current software and hardware infrastructure. 
The core behavior is meant to capture the intuition of an expert user, and to provide a level of abstraction which is easily understood by a non-expert user. 
For example, given a ground robot with no sensing, the core behaviors are Drive and Charging.
Core behaviors are similar to states in FSMs or modes in formal methods, however, core behaviors are a flexible representation at different levels of abstraction of the system, and do not require the same level of detail required in FSMs or formal methods. 
While the identification of core behavior may seem trivial, going through the steps of identifying the core behavior ensures that a baseline of performance can be established for the robot based on user intuition. 
Likewise, having core behavior identified before a transition of technology can help alleviate problems when trying to use an existing system.

For this work, we use a Capability Analysis Table (CAT) \cite{Redfield19b}, which is a documentation tool that captures how the core behaviors of the robotic system corresponds to the hardware and software of an existing autonomous system.
The CAT framework is built by a human user of the autonomous system, either during or after behavior development.
Once the CAT is developed, it can be used to capture the constraints on inputs to the system, isolate regions of failure, and provide concise representation of the system to new users.
To build and use a CAT, a spreadsheet tool is needed, along with knowledge of the inputs and expected outputs of the system (through documentation, investigation of hardware and software, and observation of system performance in the world).
CATs will not change the complexity, creation, or design of the system, but can augment documentation to improve the usability of the system. 

The main contributions to this work are: 
\begin{itemize}
    \item A method for identification of core behaviors and construction of CATs to aid in establishing a reliable robot. 
    \item A demonstration of how CATs are used to handle unreliable robot behavior. 
    \item Experimental results establishing core behavior, building CATs, and rooting out unreliable behavior for a Quadrotor.
\end{itemize}

The rest of the paper is as follows: in Section \ref{sec:relatedWork} we discussion background material for this paper. In Section \ref{sec:methodology}, we describe the methodology for identifying core behaviors and building a CAT. In Section \ref{sec:expResults}, we demonstrate the proposed methodology for an experimental platform, and in Section \ref{sec:discussion} we discussion different aspects of the proposed approach. Finally, we conclude and discuss future steps in Section \ref{sec:conclusion}.

\section{Related Work}
\label{sec:relatedWork}
There are some existing tools to provide insights into the system's core behavior, which in other areas can also be called states \cite{Chow78, Peterson77} or  modes \cite{Lyons89}. 
For example, Finite State Machines (FSMs) \cite{Lee96} are a representation of states and transitions between system states.
There are many tools that will test FSMs \cite{Chow78}, help build FSMs from state assignment \cite{Devadas88}, or represent functions as FSMs \cite{Biermann72}.
Another tool which is closely related to FSMs are Petri nets, which are a mathematical set of rules to capture distributed or asynchronous processes \cite{Murata89, Peterson77, Freedman91}.
Fernandez et al. use Petri nets as the foundation for their robotic systems \cite{Fernandez08}.
In the proposed method, they allow a human user to build a Petri net as the input commands to a robot, and provide tools to handle issues as the robot interacts with the world.
Finally, system guarantees and logical models representing a system are captured using formal methods \cite{Lyons89, DiasNeto07, Petrenko01, Bohrer19}.
More recently, Kress{-}Gazit presented a survey on how formal methods are used with robots \cite{KressGazit18} and a survey on the challenges of automatic code synthesis for reactive systems \cite{KressGazit19}. 

These tools all aid different aspects of making a reliable robot, and each captures a form of core behavior for the robot.  
FSMs capture the robot states, Petri nets capture asynchronous behavior, and formal methods represents systems logically, however the connection of the core behavior to the inputs and outputs of the system as they relate to the hardware and software of the system is missing. 
The CAT supports connections between information available to the system and the mutable variables of that system, which specifically allows for the representation of hardware.  
Additionally, CATs combine data relevant to all of these methods into a concise format, providing the user with a new representation for reasoning over autonomous system performance.

A tool to help identify how core behaviors relates to the hardware and software of the system is Systems Modeling Language (SysML).
SysML is a visual semantic framework to build models of complex systems that is an extension of Unified Modeling Language (UML) \cite{Hause06, Lenny13}, specifically, SysML breaks down systems into four components: structures, behaviors, parameters, and requirements \cite{powerpoint06}.
These models have been used to verify systems and used with different formal method approaches \cite{Laleau10, Ouchani14, Jarraya07}. 
SysML models are a powerful modeling infrastructure to model all the details of the system and require expert knowledge to build, while the CAT is a flexible methodology which is meant to benefit any user of a specific platform at certain levels of abstraction.

A similarly related area is Domain Specific Languages (DSL), which are defined as represnetations with specific vocabulary to be used by domain experts and provide some component of machine readable syntax \cite{Nordmann14}. 
Several DSLs have been built for robotics systems \cite{Gobillot14, Miyazawa19, Dhoubi12}.
In addition, a DSL has been coupled with design processes which can improve the verification and re-usability of a robotic system \cite{Schlegel10}.
Finally, a DSL has been used to map robot software to formal specification techniques. 
These approaches are modular in nature and provide improved insights on how to design and build systems, but do not focus on documentation tools for existing systems.
Likewise, these tools require the user to have domain knowledge, while CATs are intended to be a tool that can be used with limited domain knowledge. 
DSLs can take into account hardware constraints for a robotic system, but often they focus on the verification of the software. 
Because robotic systems exist with both hardware and software it is critical to start building tools which consider the failure of both components and the potential for the failure at the intersection of hardware and software. 
A survey on verification of autonomous systems presents more resources on formal methods, modeling tools, and other representations to try and assure robot behavior \cite{Luckcuck19}.
It is an active area of work to build a formal representation of CATs. 
This work is focused on a documentation tool to enable more reliable use of inherited robotic systems.

In addition to existing tools, capability analysis has been used in several fields including in statistics as the study of performance of a system \cite{Wang13} and in computing metrics for power consumption \cite{Landgren73}.
In statistics, data is taken from a system, and a variety of metrics are computed (e.g., mean, standard deviation, and range) to statistically quantify performance.
This analysis is organized in a spreadsheet of data cells called capability charts \cite{Wang13, Castagliola07}, or capability indices \cite{Kotz02}.
Alternatively, capability metrics for power consumption, are automatic tools developed by Landgren et al. \cite{Landgren73, Landgren72} to allow for faster output of performance and conditions that the network will perform under. 
Unlike the charts in capability analysis for power and system performance, the CATs discussed in this work is a tabular representation of core behavior, while connecting system inputs and outputs.

Finally, we will use CATs to perform analysis similar to root cause analysis (RCA) or fault tree analysis for a robotic system. 
RCA is the identification of what the problem is, how the problem happened, and why the problem happened, and this information is represented in maps or causal factor charts \cite{Rooney04, Dogget12}. 
Fault tree analysis is a methodology to determine the faults that may happen considering how different  combinations of software, hardware, and human inputs may disrupt the system \cite{Rauzy93, Bennetts75}.
A tool more commonly seen in robotics to generate explanations of behavior was proposed in Raman et al., who used a temporal logic approach to generating explanations of robot behavior \cite{Raman12}. 
Likewise, Hayes et al. provided explanations of robot behavior to aid in human interactions with the robot \cite{Hayes17}.
We will use CATs to provide explanations of unreliable behavior, and provide a systematic way to isolate why the undesirable behavior is occurring. 
It is more likely to see unwanted behavior when transitioning hardware and software, which is why an easy to use tool like CATs is critical to root out the negative behavior. 

\section{Methodology}
\label{sec:methodology}
\begin{figure}
    \centering
    \includegraphics[width=\columnwidth]{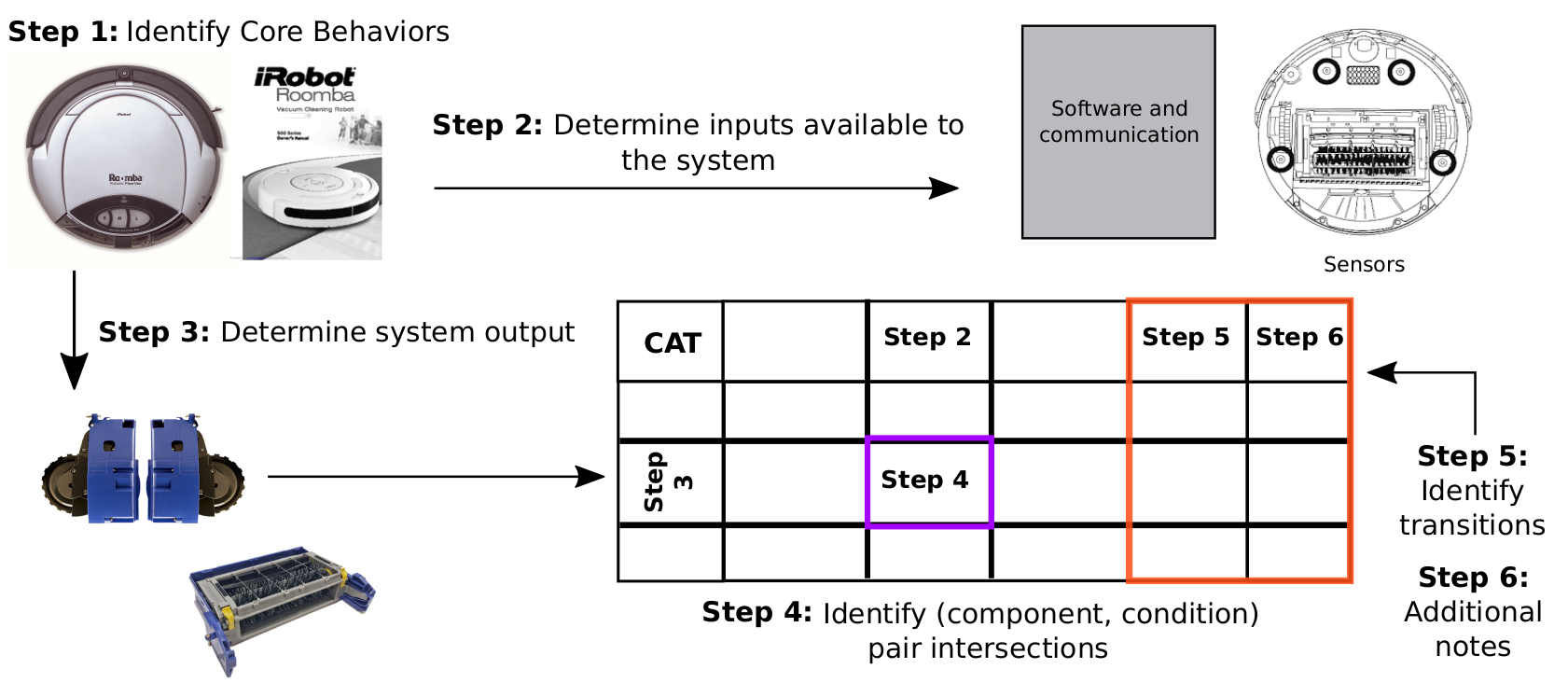}
    \caption{There are 6 main steps to building a CAT, each of which correspond to text in Section \ref{sec:buildCAT}. The Roomba images in this figure are from the iRobot website \cite{roomba}.}
    \label{fig:basicCATLayout}
\end{figure}
In this section, we will describe the identification of core behaviors, a methodology to develop CATs, and outline how CATs can be used to isolate unreliable robot behavior.
The example used throughout the Methodology will be of an iRobot Roomba \cite{roomba05}.  

\subsection{Identifying Core Behaviors} \label{sec:coreBehavior}
A core behavior is the expected behavior of the system from basic inspection, without a need for large amounts of detail about how the behavior is implemented. 
Identifying core behaviors does not require expert knowledge of the system, and as the users understanding of the system evolves the core behaviors can be refined. 
Core behaviors for some robots may be obvious, and in other cases may require a more in depth exploration of the system or consultation with an expert. 
Sources of information which can help shed light on the core behaviors are: code documentation, hardware manuals, inspection of the code and hardware components, consulting an expert, and, when necessary, experimental trials. 
Of these options, running experimental trials is the most dangerous, especially when proper inputs and outputs are not identified for the system. 
For the Roomba example, we used a specification manual to identify the core behaviors \cite{roomba05}. 

Core behavior is identified by understanding the basic functionalities of the robot. 
Questions should be asked, such as: \textit{If the robot is turned on, what will the robot do? Given this input what output is expected?}
Identifying core behaviors captures the users intuition of what the robot should do, and allows for a standard representation. 
Additionally, core behaviors abstract away the need to have intimate knowledge of the robot's subsystems. 
The key at this level of abstraction is to select behaviors which highlight the functions that must persist for the robot to do anything meaningful in the world. 
The early generation the Roomba has 3 core behaviors: Drive, Clean, and Charging. 
These are the three basic states of the robot's expected overall behavior. 

Once identified, a set of baseline experiments can be established to ensure the persistence of the core behavior.  
In doing so, the robot has a performance benchmark which can be tested and evaluated before more advanced capabilities are added. 
This can enable principled experimentation of new algorithms and more reliable robot behavior.
Establishing this baseline is critical to knowing if the system is performing reliably, and can be used throughout the lifespan of the robot. 

For the Roomba example, a set of baseline experiments which evaluate the core behaviors are: 
\begin{itemize}
    \item the robot increases power and does not move when charging,
    \item the robot drives with no warning light,
    \item the robot vacuums with no warning light.
\end{itemize}

Each of these experiments evaluates specific underlying components of the core behaviors which we will identify in Section \ref{sec:buildCAT}

\subsection{Building a Capability Analysis Table} \label{sec:buildCAT}
In this section, we will define a methodology for how to build a Capability Analysis Table (CAT) for an existing system.
The same information used in the identification of the core behaviors can be used to build a CAT. 
Figure \ref{fig:basicCATLayout} is a basic outline of a CAT.
For more details on how to build a CAT for design purposes, the reader is directed to \cite{Redfield19b}.

\begin{figure}
  \centering
  \includegraphics[width=\columnwidth]{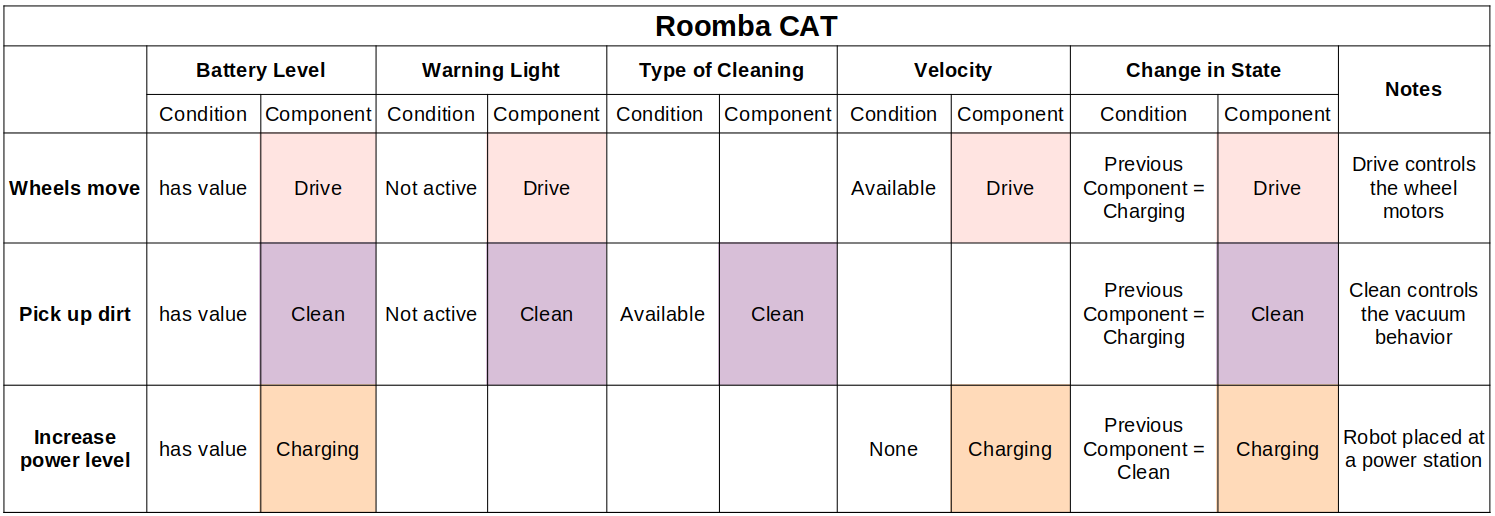}
  \caption{Given an early iRobot Roomba system based on information from \cite{roomba05}, we built a CAT which describes the robot's core behavior. The three core robot behaviors are: Drive, Clean, and Charging. }
  \label{fig:roombaExample}
\end{figure}

\textbf{Step 1} in building a CAT is to identify the core behaviors of the system, outlined in Section \ref{sec:coreBehavior}.
We will build an example CAT for the Roomba, Figure \ref{fig:roombaExample}.
The Roomba core behaviors were identified to be Drive, Clean, and Charging.

\textbf{Step 2} is to identify the inputs to the system, which includes both inputs from sensors, inputs generated within software, and communication which is provided to the robot.
These inputs make up the column headers, and are named by the user. 
Consistency and clarity in naming will aid in future use.
The main goal is a documentation aid which incorporates both hardware and software constraints.
The Roomba has a variety of inputs, but the key required inputs for the core behaviors defined above are: battery level, warning light, type of cleaning, and velocity. 
Note that many more inputs might be available, but at this level of abstraction all of these inputs are necessary in some way for the core behaviors to exist.

\textbf{Step 3} is to identify the expected outputs or actions that are expected at the core behavior level of the system.
These outputs make up the left-hand side row headers.
In systems which have both hardware and software components, it is important to consider how outputs are represented in software for the different core behaviors. 
For the Roomba example: Drive has an output of wheels moving, Clean has an output of pick up dirt, and Charging has an output of increasing the power level. 

\textbf{Step 4} is taking the defined inputs and outputs that have been defined and filling in the cells of the CAT.
To do so, start with considering a corresponding output and which core behavior contributes to that output.
Each column is composed of a (Condition, Component) pair, which represent the constraints the input may require for the specific core behavior.
Along each column put the core behavior in the Component grid cell, and the corresponding constraint in the Condition grid cell. 
For each row, every column with an entry must meet the specified condition to see the desired output. 
Blank spaces along a row represent inputs that are not necessary for the core behavior to generate the desired output.
CATs are a tabular representation of a set of conditions which can be represented logically. 

For example consider the Roomba CAT in Figure \ref{fig:roombaExample}.
Let us use the core behavior Drive and the output wheels move.
Along the row, Drive will be placed in the Component grid cell for the following columns: Battery level, warning light, and velocity.
This implies that for the output, wheelsMove, all of these conditions must be true. 
This can be represented logically as: 

\begin{equation}
   \mathrm{Drive} = (\mathrm{batteryLevel} \land \mathrm{warningLight} \land \mathrm{velocity} \land \mathrm{changeInState} ) \implies \mathrm{wheelsMove}
   \label{eq:math}
\end{equation}

This type of logical representation can be built for every row, where all the input columns that intersect the row with an entry are a component, condition pair which is required by a logical AND condition for the output to exist. 
To achieve a logical OR statement, one output will have two or more sub-rows, which is seen in the more complex CAT in Figure \ref{fig:pelicanCAT}. 
Each column with a component entry along a sub row will be combined using the logical AND as done in Equation \ref{eq:math}. 
The resulting sub row statements will be combined for a single output using the logical OR.  

\textbf{Step 5} is to identify the conditions which will transition the system into the core behavior for that row, similar to a tabular FSM. 
Transitions between states may be dependent on time or it may be possible to have asynchronous execution of behaviors. 
Timing constraints can be handled in the condition sub-column, and it is a direction of future work to consider how CATs can be used in real time incorporating these complexities. 

\textbf{Step 6} is the final step, and it is the process of adding notes to the final column to provide any additional information to the user.  
For the Roomba example, the CAT in Figure \ref{fig:roombaExample} is filled in to reflect the transitions between behaviors and a discussion of each core behavior.
Once built, CATs can be used as a documentation aid to help transition robot capabilities to new people, maintain up to date system information, and provide tools to troubleshoot negative robot behavior.

\subsubsection{Handling Levels of Abstraction}
As robotic systems grow in complexity, CATs can establish different levels of abstraction, highlighting different levels of detail. 
The more CATs built the more details available about the relationships between the hardware and software of the system. 
The ability to abstract at different levels simplifies the raw information that must be represented at any one level, and improves the users ability to use the system reliably.

\begin{figure}
\centering
  \includegraphics[width=\columnwidth]{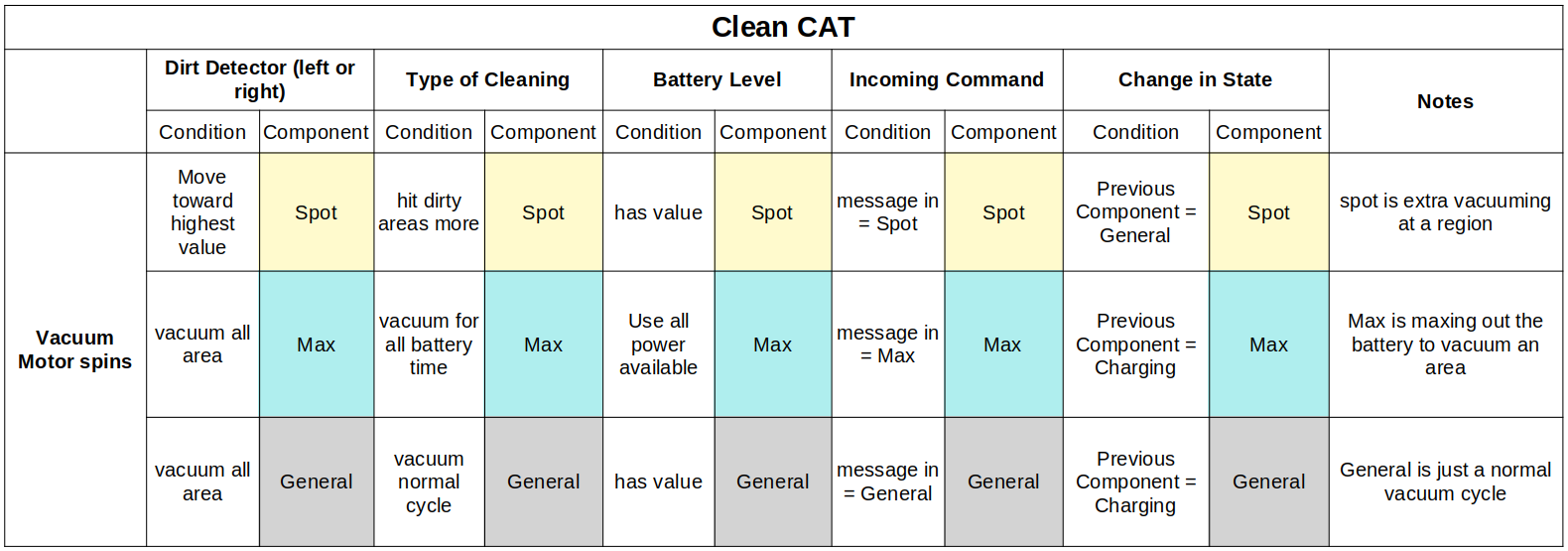}
   \caption{The Clean Roomba core behavior decomposed into a CAT. }
   \label{fig:roombaClean}
\end{figure}

Consider that more detail is needed about a core behavior.
The abstraction can be decomposed into lower level robot behaviors using the steps outlined in Section \ref{sec:buildCAT}.
The first step is to select the core behavior to be further decomposed.
Next, for that behavior, a set of new core behaviors which are at a lower level of abstraction are selected.
Finally, using the steps in Section \ref{sec:buildCAT} a lower level CAT is built. 

\begin{figure}
    \includegraphics[width=\columnwidth]{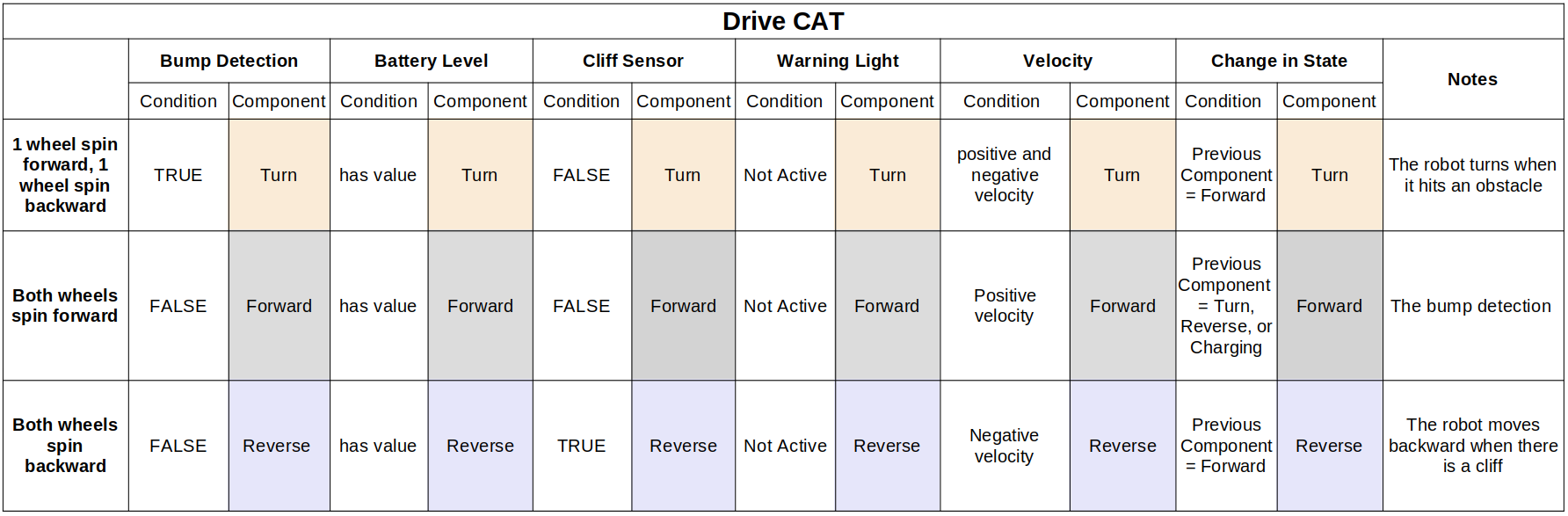}
    \caption{The Drive Roomba core behavior decomposed into a CAT.}
  \label{fig:roombaDrive}
\end{figure}

Using the Roomba example, we select Drive and Clean as behaviors which are composed of more detailed core behaviors. 
First, we decompose Clean into these new core behaviors: Spot, Max, and General. 
These key behaviors are types of Vacuuming protocol for the Roomba. 
The output of these behaviors is to have the vacuum motor spin and depending on different incoming commands, and the amount of dirt detected, different core behaviors will be expressed. 
This lower level CAT can be seen in Figure \ref{fig:roombaClean}. 

Next, consider Drive as a core behavior. 
This can be broken down into three new core behaviors: Turn, Forward, and Reverse. 
The output of these new behaviors are different combinations of wheels turning and the inputs are based on different sensor values. 
For example, to Turn the bump detection sensor must be true, and the cliff sensor must be false in addition to having battery and a velocity value, as soon as the condition of bump detection is false the system can transition back to the forward component. 
The same reasoning can be worked through for each row of Figure \ref{fig:roombaDrive}

\begin{figure}
  \centering
  \includegraphics[width=\columnwidth]{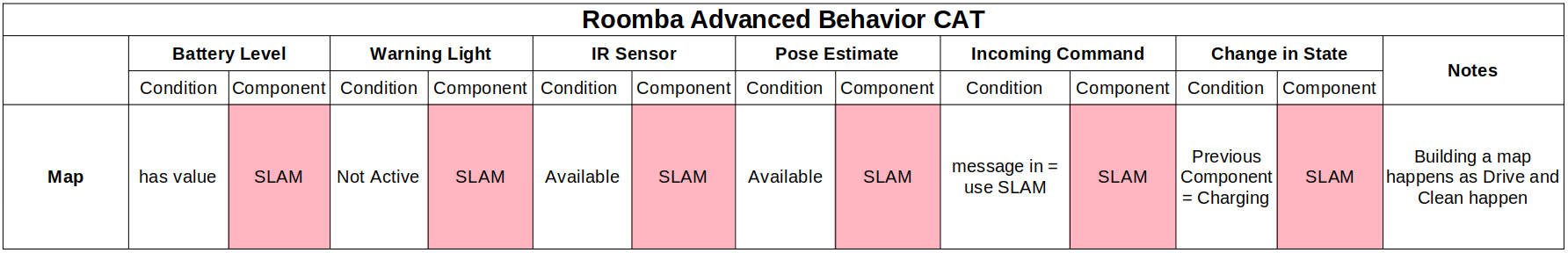}
  \caption{To update a CAT either modify an existing CAT or build a new CAT which can sit on-top of or in parallel with core behavior already established. This CAT is for the new SLAM Roomba feature and captures the necessary sensor data and has a map as the resulting output.}
  \label{fig:roombaExampleHighLevel}
\end{figure}
In the life cycle of a system, it is possible to have many updates to the hardware and software components.
It is important to keep the CAT up to date for it to remain useful to future users,.
Likewise, CATs can be built to inherit properties of lower level CATs as more complex behaviors are added. 

For example, in newer iterations of the iRobot Roomba, Simultaneous Localization and Mapping (SLAM) occurs on board the robot. 
To add this feature to the CAT, we make a Roomba Advanced Behavior CAT which has SLAM as a core behavior and map as an output. 
The necessary inputs are battery level, warning light, IR sensor, pose estimate, and incoming command. 
This CAT can be seen in Figure \ref{fig:roombaExampleHighLevel}.
While SLAM is a well understood capability, this CAT is able to abstract away code and specific hardware details, and capture the necessary components to communicate to the user how certain outputs are achieved. 
The CAT in Figure \ref{fig:roombaExampleHighLevel} was built separate of the core behavior CAT, however, in this case, the added complex behavior does not inherit properties from the core behavior CAT and could have been a new row in the original CAT, Figure \ref{fig:roombaExample}. 
It is possible that the IR sensor is also used in one of the original core behaviors. 
To add a new sensor to the Roomba CAT, Figure \ref{fig:roombaExample}, append a new column to the table for the new sensor, and update the entries along each row for core behavior that use the new information.

\subsection{Rooting out Unreliability} \label{sec:debugCAT}
Failures are going to occur in the lifetime of an autonomous system.
The key is being able to identify the failure points, and capture if the issue is human error, environmentally caused, or hardware/software related. 

First, in cases where a failure has occurred, identify which core behavior was violated.
Once a core behavior has been isolated, consider if it is clear that a condition was broken at the highest level CAT.
If a condition at this level is the culprit, then resolve the condition for that component and move on.
If the behavior was violated but a condition was not broken, consult lower level CATs to determine which lower level behavior condition was violated.
This process can be done iteratively until the broken condition is discovered.

For example, if while the Roomba is in the Drive core behavior the robot stops and the battery is below the threshold constraint then the robot should transition to the Charging behavior. 
This is a state transition whose trigger is represented in the highest level CAT and should be resolved at that level. 
However, if the robot stops and the warning light is active then lower level CATs need to be consulted, examples of lower level CATs are in Figure \ref{fig:roombaClean} and Figure \ref{fig:roombaDrive}

It is possible for the CAT to be incorrectly constructed because this is a human generated tool. 
As errors are found and more knowledge is gained about the system, the CAT can be easily fixed for future by updating the entries. 

\section{Experiment with Quadrotor} 
\label{sec:expResults}
\begin{figure}
  \centering
  \includegraphics[width=0.75\columnwidth]{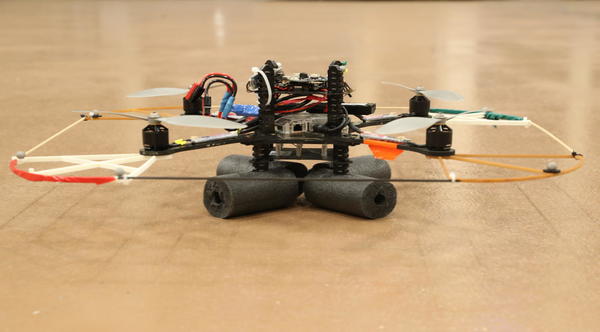}
  \caption{An AscTec Pelican Quadrotor}
  \label{fig:pelican}
\end{figure}

We identified core behavior, established baseline experiments, built CATs, and isolated failure cases for the Ascending Technologies Inc. Pelican Quadrotor (AscTec Pelican Quadrotor), Figure \ref{fig:pelican}.
New users inherited this robotic system with an existing software infrastructure, and it was necessary in a short amount of time to have the robot operational. 

\subsection{Identifying Core Behaviors and Baseline Tests}\label{sec:expCore}
\begin{table}
  \centering
  \begin{tabular}{|c|c|}
    \hline
    Number of Launch Files & $147$ \\ \hline
    Lines of Code & $37,036$ \\ \hline
    Hardware Pieces & $>20$\\ \hline
  \end{tabular}
  \caption{Parts making up the AscTec Pelican Quadrotor}
  \label{table:robotComponents}
\end{table}

To identify the core behaviors of the AscTec Pelican quadrotor and the accompanying inherited software, we went through the provided information outlined in Table \ref{table:robotComponents}.
This information was gathered through the inspection of: hardware manuals, software documentation, and physical attributes.
In Table \ref{table:robotComponents}, the number of launch files corresponds to all files with a .launch file type within the repository for the history of the robot.
This repository has had many managers and has been in operation for over 10 years, collecting significant code rot over time.
Likewise, the total number of lines of code for the system, shown in Table \ref{table:robotComponents}, consists of both on board robot firmware and ground station code written in: C++, embedded C, and python. 
Note that not all of this code is used to control the robot, and upon first inspection it was unclear how the software connected to the hardware to allow the robot to fly. 
Finally, in Table \ref{table:robotComponents}, the robot hardware consists of 4 motors, 4 Electronic Speed Controllers (ESCs), 4 propellers, 1 low level autopilot, 1 high level autopilot, sensors on both autopilots, connections between autopilots, connection from autopilot to a high level on board computer, WiFi dongle, and the frame of the robot. 
The hardware components were determined with physical inspection, and is a rough estimate of components where some parts like sensors on the autopilot and the frame are grouped together, and things like motors and ESCs are listed out. 
For a more detailed estimate, hardware manuals needed to be consulted.
Note that counting out all of the sensors on the autopilot and the individual frame components would only increase overall number of robot components, increasing the overall complexity of the system. 
Table \ref{table:robotComponents} highlights that a robot with relatively simple capabilities is comprised of many raw components.
However, the basic intuition for what the robot should be able to do is: Takeoff, Fly, and Land. 
With further inspection, and some trial and error experimentation (with high damage cost to the robot) we realized that Idle and Emergency were also necessary core behaviors. 

The establishment of core behaviors allowed for the development of a set of experiments to establish baseline performance of the robot. 
We used two experiments: a ``hover'' experiment and a ``square'' experiment.
In the hover experiment, the robot took off, hovered at 1\,m for 5 seconds, and then hovered at 2\,m for 5 seconds and landed.
In the square experiment, the robot flew in a 1.5m square at a fixed altitude. 
Both test cases contributed to testing the core behavior Fly along with transitions between different core behaviors like Takeoff to Fly to Land. 

\subsection{Building a Capability Analysis Table}
\begin{figure}
  \centering
  \includegraphics[width=\columnwidth]{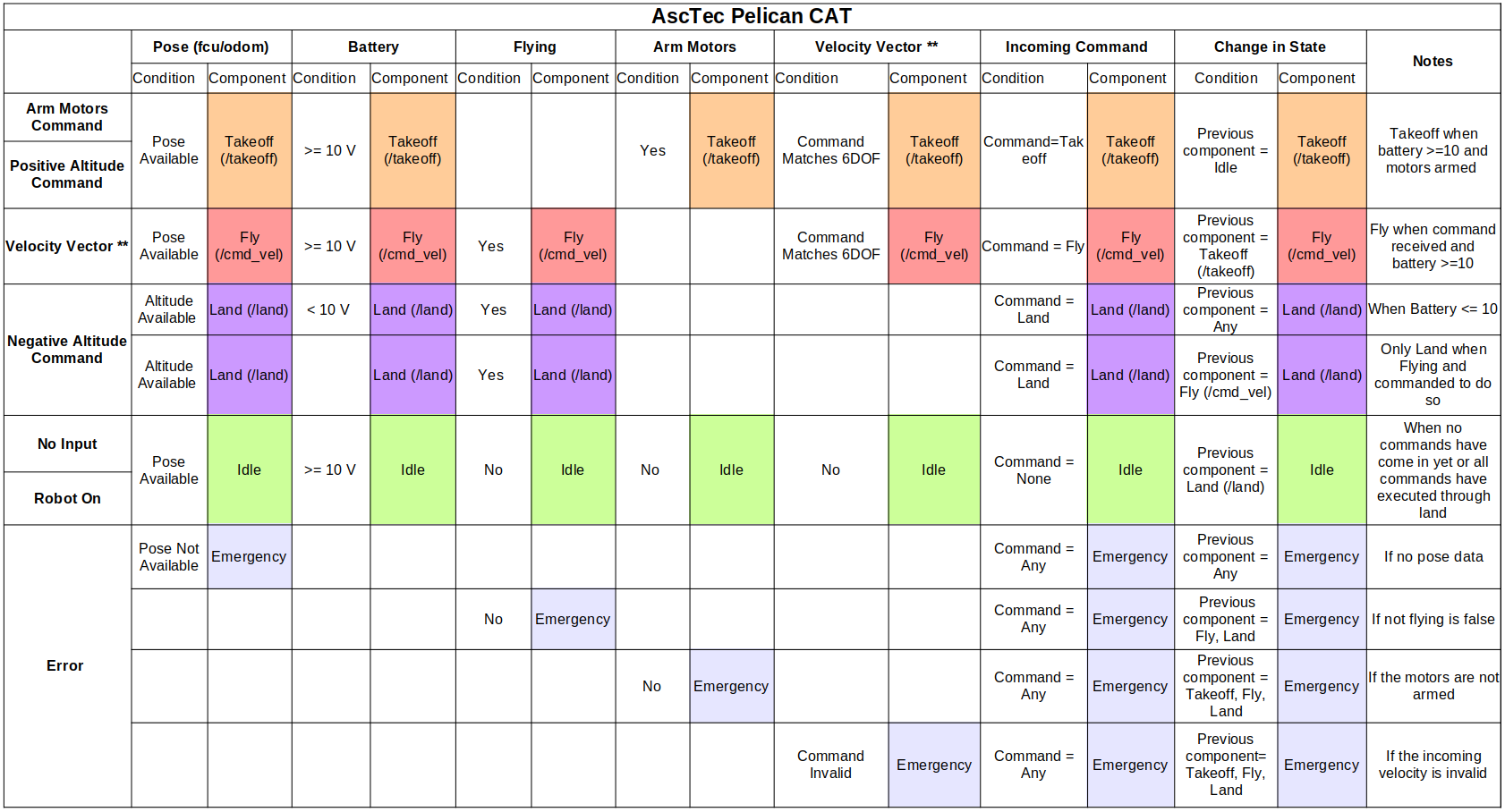}
  \caption{A CAT built for the AscTec Pelican Quadrotor at the core behavior level of abstraction.}
  \label{fig:pelicanCAT}
\end{figure}

\begin{figure}
  \centering
  \includegraphics[width=\columnwidth]{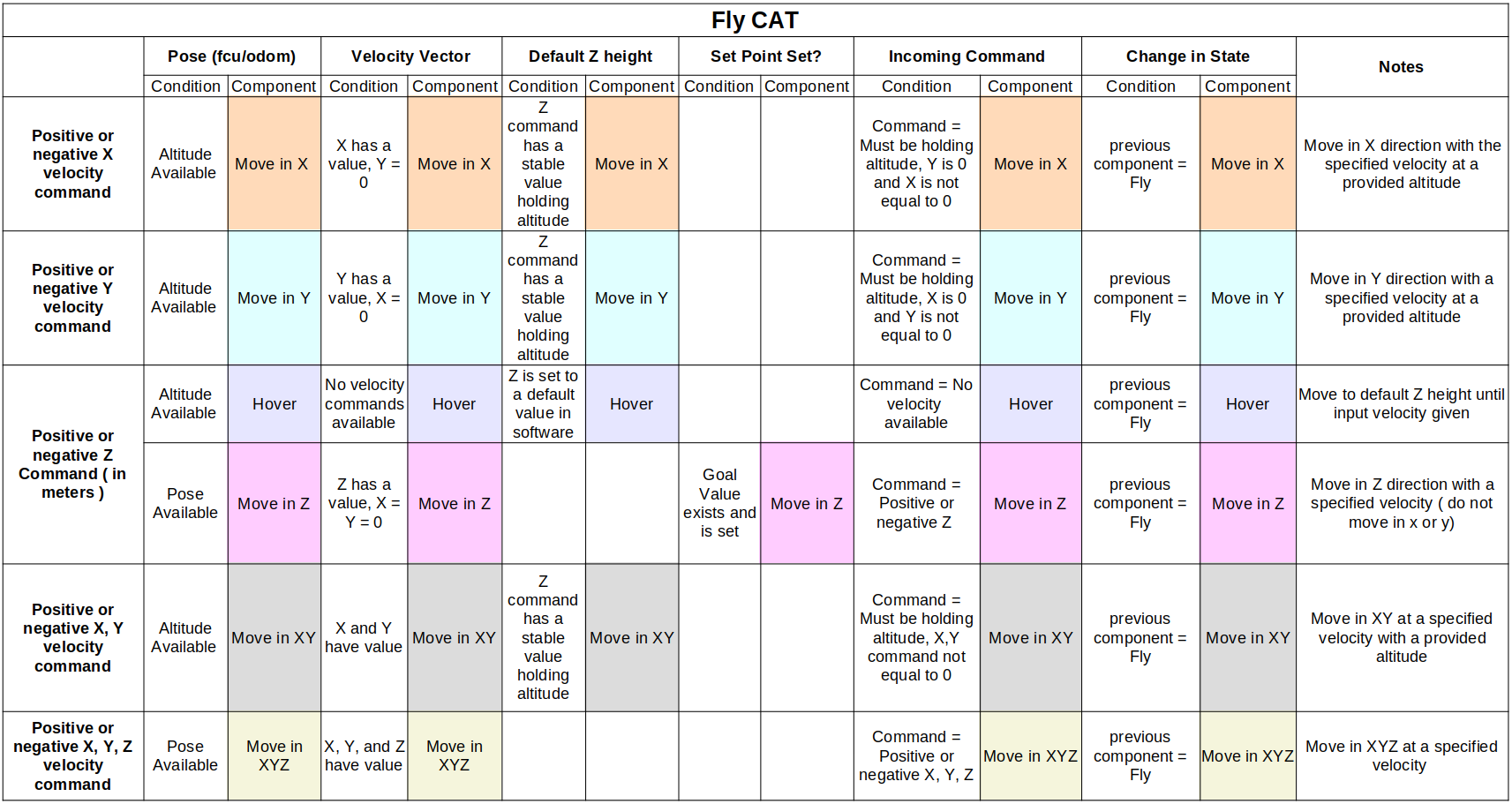}
  \caption{A CAT for the Fly core behavior. This CAT breaks down flying into different core behavior which provide details on different combinations of movement the robot can execute.}
  \label{fig:pelicanFlyCAT}
\end{figure}

Following the steps in Section \ref{sec:buildCAT}, we built a CAT for the AscTec Pelican Quadrotor's identified core behaviors:

\begin{enumerate}
\item Identify core behaviors: Takeoff, Land, Fly, Idle, Emergency. Develop a set of test cases to evaluate core behaviors: hover experiment and square experiment. 
\item Identify available inputs (sensors, communication, or inputs) to the system: Pose, battery, flying, arm motors, velocity vector.
\item Identify expected outputs or actions of the system: Velocity commands in positive or negative Z, Velocity commands, no input, robot on, arm motors, error.
\item Example of entering behaviors across the row: Takeoff you have outputs of arm motors and positive altitude, which requires input from pose, batter, arm motors, and velocity vectors. In each Condition, Component pair, the constraint on the input is listed in the Condition column and the core behavior is listed in the Component column.  Flying is left blank because this input is not necessary for this behavior. The remaining CAT entries can be seen in Figure \ref{fig:pelicanCAT}.
\item Complete the Change in Component column with the transitions between different components.
\item Finally add notes in the Notes column to describe the component and any conditions on that row. 
\item To break the CAT in Figure \ref{fig:pelicanCAT} into sub components, consider the Fly behavior. We will repeat the procedure outlined in Section \ref{sec:buildCAT} to construct a more detailed description of the Fly behavior, this new CAT is in Figure \ref{fig:pelicanFlyCAT}.
\end{enumerate}

Table \ref{table:catComponents} outlines the number of elements that make up the CATs for the AscTec Pelican Quadrotor. 
The two CATs built are a concise representation of all of the raw information in Table \ref{table:robotComponents}.
The CAT reduced a large amount of raw data into a clear format representing the core behavior of the robot to allow for test and evaluation of key properties.
Now that the CATs exist, they can be used as additional documentation for the robot's core behavior for future users to work with the robots reliably. 

\begin{table}
  \centering
  \begin{tabular}{|c|c|}
    \hline
    Number of Core Behaviors & 11\\ \hline
    Number of Outputs & 12\\ \hline
    Number of Inputs & 13\\ \hline
    Number of (Condition, Component) pairs & 70\\ \hline
  \end{tabular}
  \caption{Details on CAT components}
    \label{table:catComponents}
\end{table}

\subsection{Evaluation}
  \begin{figure}
    \centering
    \subfigure[Hover Experiment Failure]{
    \includegraphics[width=0.45\columnwidth]{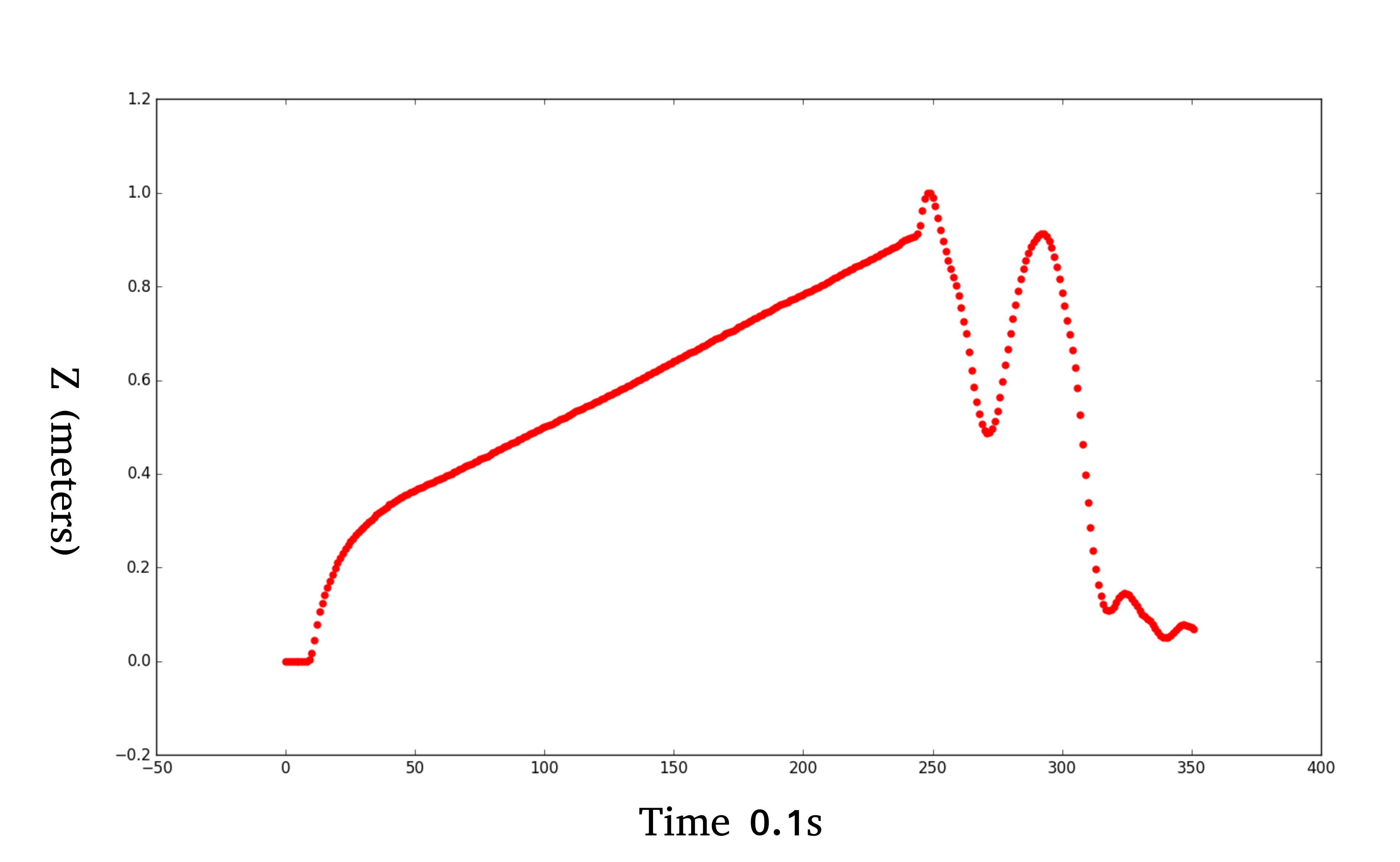}
        \label{fig:hoverExpWrong}
    }
    \subfigure[Hover Experiment Success]{
        \includegraphics[width=0.45\columnwidth]{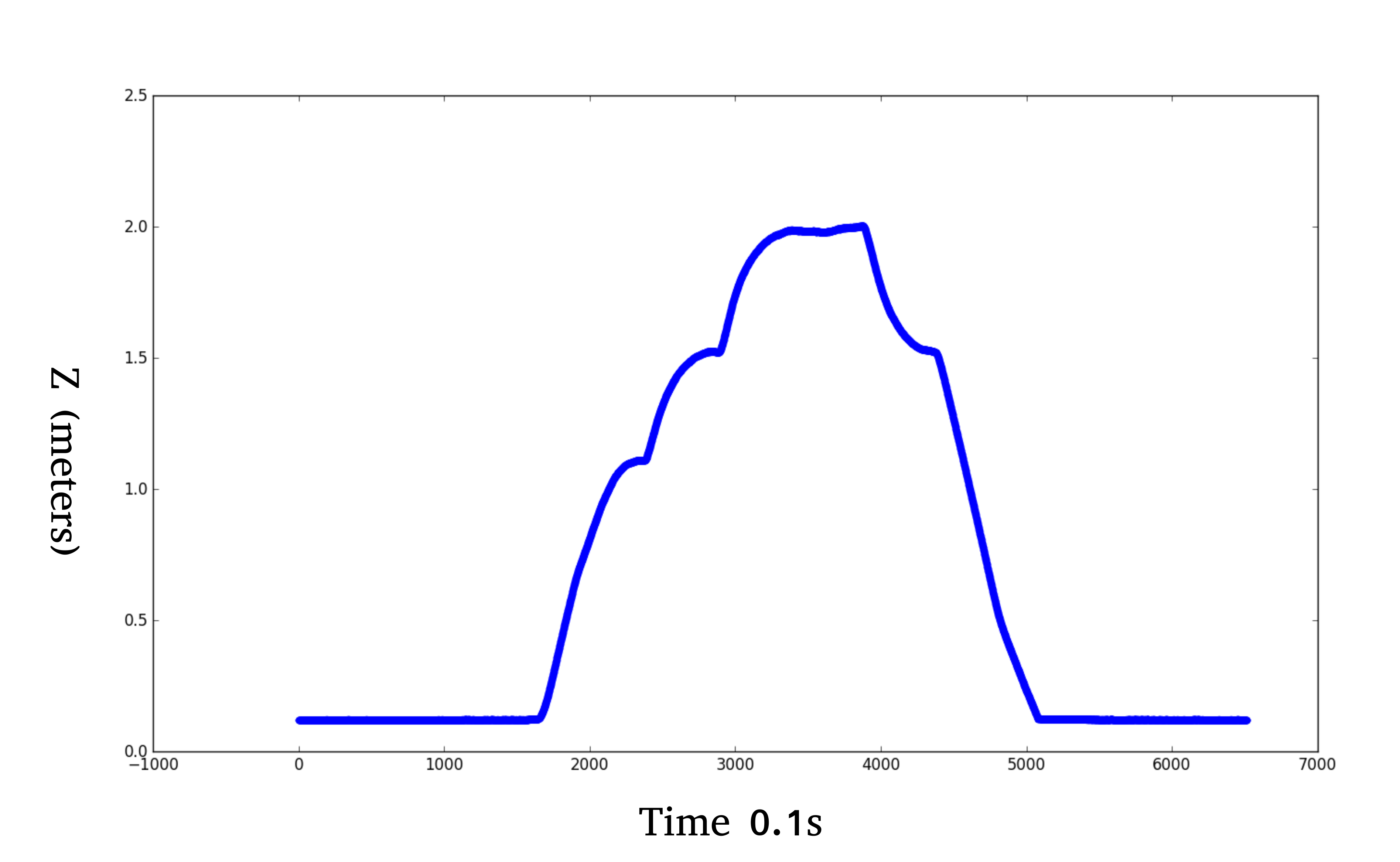}
        \label{fig:hoverExpCorrect}}
    \subfigure[Square Experiment 3D view]{
        \includegraphics[width=0.45\columnwidth]{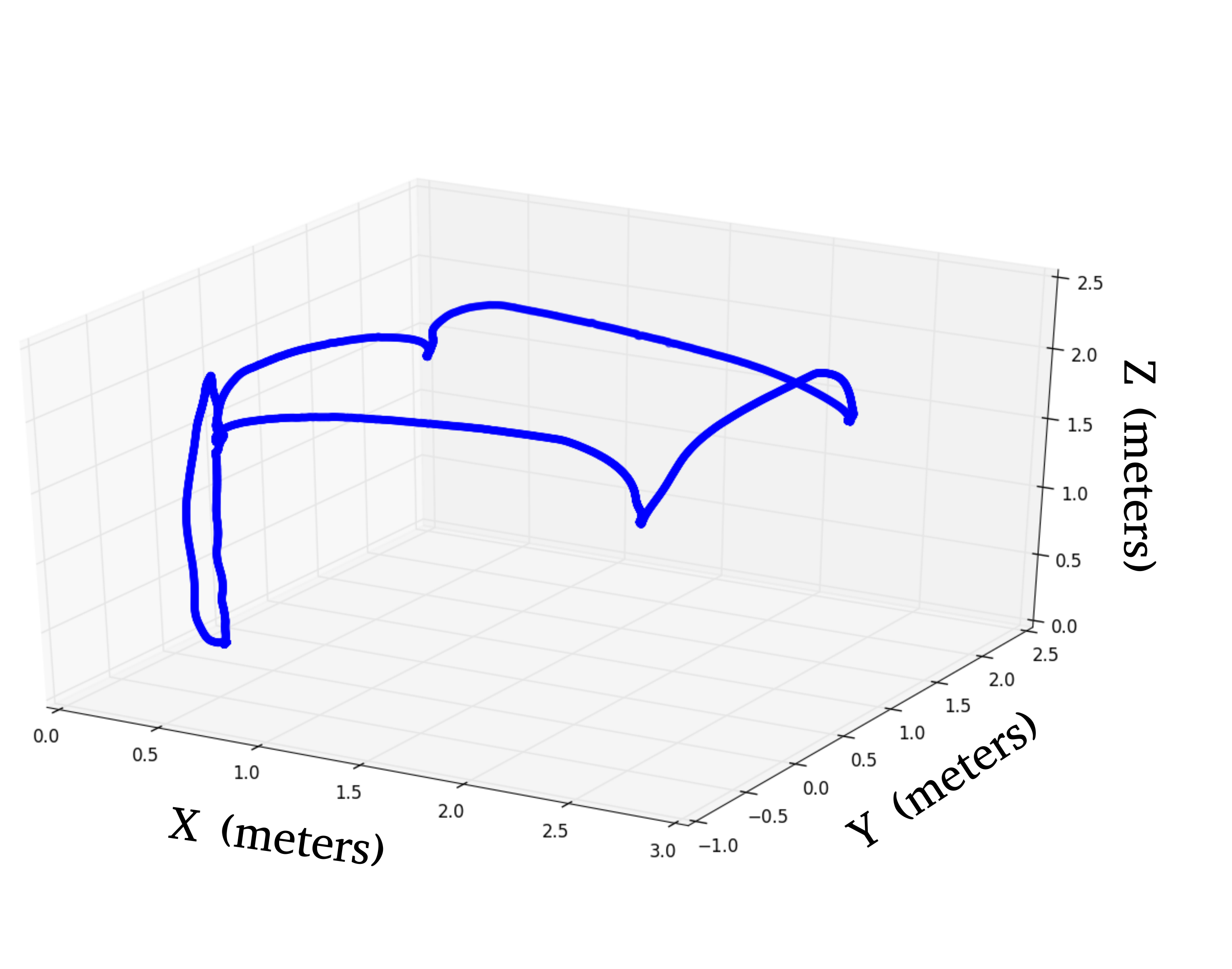}
    \label{fig:squareExp3D}
    }
    \subfigure[Square Experiment 2D view]{
        \includegraphics[width=0.45\columnwidth]{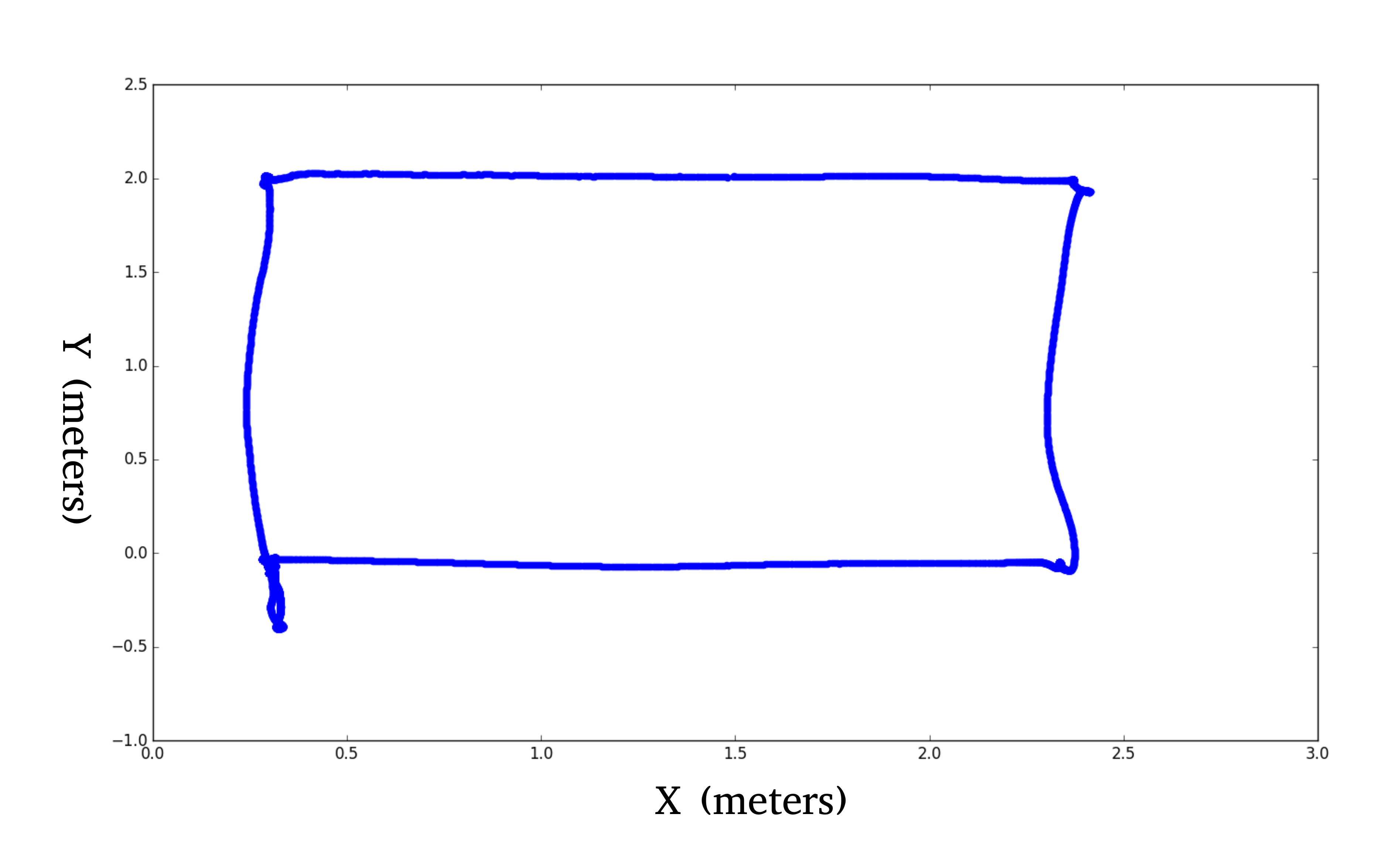}
    \label{fig:squareExp2D}
    }
  
    \caption{
    Figure \ref{fig:hoverExpWrong} shows the z height in meters of an unsuccessful hover experiment where the robot took off and drifted toward the ceiling. 
    Figure \ref{fig:hoverExpCorrect} shows the resulting z height in meters for a successful hover experiment.
    Figure \ref{fig:squareExp3D} shows the 3D path of the robot through space during a square experiment. 
    Figure \ref{fig:squareExp2D} is the 2D projection of the robots path through space. }
    \label{experimentalResults}
    \end{figure}

The proposed benchmark experiments in Section \ref{sec:expCore} were run to see that the robot was performing core behavior as expected, Figure \ref{experimentalResults}.
These tests where important to run to establish that the robot behaved reliably, especially, after a long period between experiments.  
Figure \ref{fig:hoverExpCorrect} shows the completed trajectory of a successful hover experiment, Figure \ref{fig:squareExp3D} shows the completed 3D trajectory of the square experiment and Figure \ref{fig:squareExp2D} shows the 2D projection of the 3D trajectory completed in the square experiment.
These tests ensure that UAV the core behaviors are reliable before adding more complex algorithms or putting the robot in an environment with greater uncertainty. 

\subsection{Repair and Redesign} 
When first using the AscTec Pelican quadrotor several problems were encountered: hardware, software, and human input error. 
For example, after takeoff, the robot would occasionally drift toward the ceiling.
Figure \ref{fig:hoverExpWrong} shows an example of this happening where the robot drifted toward the ceiling until the safety pilot took over at $\approx 23.0$\,s.

The steps in Section \ref{sec:debugCAT} where followed to identify the core behavior that was failing: in this case it was the Fly behavior. 
The CAT in Figure \ref{fig:pelicanCAT} was inspected and after ensuring that all conditions for the FLY behavior were met, we went to the CAT in Figure \ref{fig:pelicanFlyCAT} which breaks down the Fly core behavior. 
Within this CAT we identified Hover as the core behavior that was failing. 
It was determined that if no velocity commands were provided, and a default altitude was not set the robot would drift toward the ceiling. 
Thus a condition for Hover was broken, resulting in the unexpected behavior. 
This problem was resolved, and a successful Hover experiment is shown in Figure \ref{fig:hoverExpCorrect}.

It is possible for the CAT to be incorrectly built, especially if the person is a new user and building the CAT while learning the system. 
The CAT provides a framework to systematically isolate the users intuition as it relates to what is known about the hardware and software. 
If the CAT is incorrect, it is possible to fix the table by further investigating the raw information which makes up the robot.
Over time and with increased understanding of the system, the CAT will become more complete, descriptive, and provide meaningful information to future users.

\section{Discussion}
\label{sec:discussion}
Building a CAT is more easily done by a person that has intimate knowledge of how a robot works because there is less need to consult documentation or other sources of information about the system.   
However, non-expert robot users can build a CAT by using basic intuition and inspecting the raw information about the system.
For example, the Roomba core behavior CAT, Figure \ref{fig:roombaExample}, is able to capture the core behavior of the system with limited technical details.
The only information available was an online specification \cite{roomba05}. 

CATs have the power to condense large amounts of raw information and grow flexibly with additional capabilities. 
The complexity of CATs will vary system to system, but the abstraction allows CATs to describe the robotic system at the necessary level of detail for the documentation. 
Developing CATs may take time, but investing in the documentation of hardware and software relationships can ease burdens on future users. 
Likewise, building a CAT provides a method which can translate intuition into: core behaviors, a set of benchmark tests, and a way to eliminate unreliable behavior. 

An added advantage of CATs is that they can also help provide clear explanations for errors which may occur in the system. 
New users face errors that experienced users may not encounter due to years of developing intuition for how robotic systems work. 
A CAT can reduce the barrier to entry to allow new users to trouble shoot problems before hitting a wall or needing more expert assistance. 

Finally, the identification of core behaviors at the correct level of abstraction allow for benchmark performance evaluations to be established. 
Having such performance measures when changing to a more complex environment or adding new capabilities will provide insight into what the robot should do, and if the robot is physically capable of doing the task in the new environment. 
For example, consider a Huskey robot with a 7DOF robot arm attached. 
Both the Husky ground robot and the 7DOF robot arm will have individual CATs which capture the core behavior of the separate capabilities. 
A new table is necessary to now capture that the robot can both move and grasp simultaneously, expanding what the set of core behaviors of the robot.

\section{Conclusion and Future Work}
\label{sec:conclusion}
To establish reliable robot behavior we used Capability Analysis Tables to connect the robot's core behaviors with the necessary hardware and software inputs and outputs of the system.
Unlike existing methods, CATs connect raw information about the robotic system's core behaviors ensuring: more robust evaluation of future algorithms, an established set of requirements, and a way to root out unreliable behavior.
We use CATs to address the need for user friendly documentation to capture the ever growing complexity of autonomous systems. 
In providing users with a more concise representation of the autonomous system, more reliable behavior can be established. 

Future work will consider how a core set of behaviors might span across many vehicle types. 
For example, do all UAVs share a set of core of behaviors? 
Is a potential set: Takeoff, Fly, Land, Emergency, and Idle? 
It is our hypothesis that this generalization can be made for different robot vehicle types, and that the combination of hybrid vehicle types will require additional CATs to illuminate the new behavior achieved.
The specific hardware and software details for each robot would differ from platform to platform, but establishing a set of core robot behaviors could allow for a ubiquitous performance evaluation for robots in a certain category. 

In addition, it is a direction of future work to make CATs machine readable and perform automatic updates to the table, along with consistency checking. 
The issue remains of addressing keeping CATs human readable, and not imposing syntax that turns the tool into something like DSL or SysML which require advanced knowledge of the specification methods to use. 
There is active work considering how to map CATs to existing formal methods tools which will help ensure the validity of the current table structure. 
We are also working on how a CAT could be coupled with robot logs to do run time monitoring of the system.
These directions will continue to improve the CAT methodology and evolve the tool past a documentation aid for ensuring reliable robot behavior. 

\nocite{*}
\bibliographystyle{eptcs}
\bibliography{CATPaper}

\end{document}